\documentclass{article}
\usepackage[table]{xcolor}

\usepackage[preprint]{corl_2026} 
\usepackage{float}
\usepackage{graphicx}
\usepackage{amsmath,amssymb}
\usepackage{booktabs}
\usepackage{caption}
\usepackage{makecell}
\newcommand{\best}[1]{\cellcolor{orange!25}\textbf{#1}}

\title{HOIST: Humanoid Optimization with Imitation and Sample-efficient Tuning for Manipulating Suspended Loads}

\author{
  Songyang Liu\thanks{Equal contribution.}\\
  Department of Civil and Coastal Engineering \\
  University of Florida 
  United States\\
  \texttt{liusongyang@ufl.edu} \\
  \And
  Shunyu Yao\footnotemark[1] \\
  Department of Civil and Coastal Engineering \\
  University of Florida 
  United States\\
  \texttt{shunyu.yao@ufl.edu} \\
  \AND
  Dingyuan Huang \\
  Department of Civil and Coastal Engineering \\
  University of Florida 
  United States\\
  \texttt{dingyuanhuang@ufl.edu} \\
  \And
  Shuai Li \\
  Department of Civil and Coastal Engineering \\
  University of Florida 
  United States\\
  \texttt{shuai.li@ufl.edu} \\
  $^{\dagger}$Corresponding author
}

\begin{document}
\maketitle 


\begin{abstract}
Manipulating suspended payloads with humanoid robots is challenging because the robot can only influence an underactuated, oscillatory load through whole-body motion and intermittent contact. 
Imitation learning provides safe initial behavior but does not directly optimize final placement, while reinforcement learning from scratch is unsafe and sample-inefficient on real humanoids. 
We present \textsc{HOIST}--\textbf{H}umanoid~\textbf{O}ptimized with~\textbf{I}mitation and \textbf{S}ample-efficient \textbf{T}uning for manipulating suspended loads. 
\textsc{HOIST} first finetunes a high-level vision-language-action (VLA) policy from virtual-reality (VR) teleoperation demonstrations and executes its commands through a whole-body controller. 
It then uses VLA rollouts and iterative batched RL to improve placement accuracy and stopping behavior. 
Experiments in simulation and on a real humanoid show that \textsc{HOIST} improves over imitation-only and additional-demonstration baselines; compared with pure VLA rollouts, \textsc{HOIST} reduces translational placement error by 19.9~$cm$ and raw angular error by 3.56$^\circ$, demonstrating the potential of humanoids for underactuated material-handling tasks.
\end{abstract}





\begin{center}
    \vspace{-0.8em}
    \includegraphics[width=0.7\linewidth]{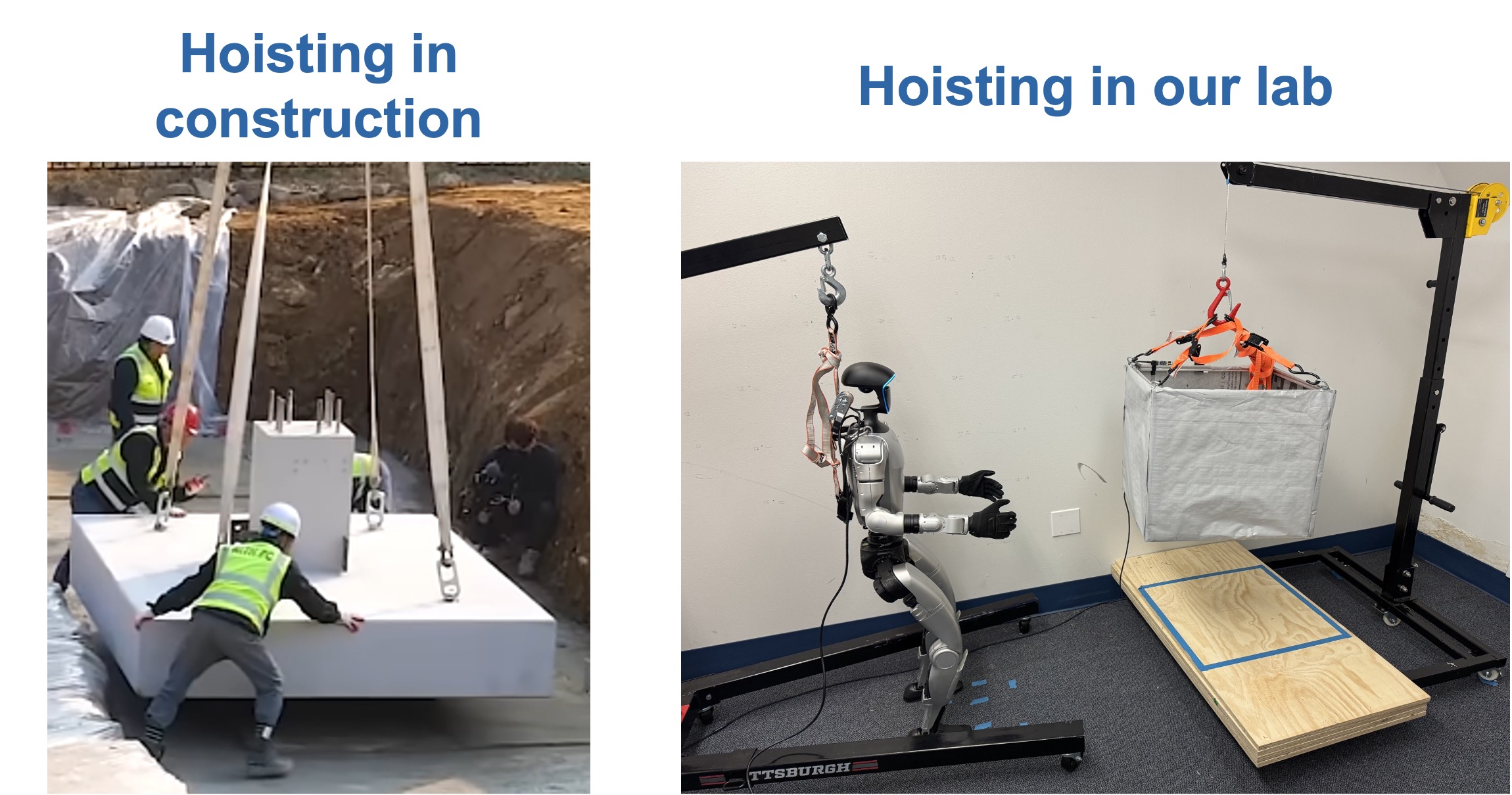}
    \captionof{figure}{
    Motivation and task setup for humanoid hoisting. 
    Left: workers manually guide a suspended payload during construction hoisting, exposing them to struck-by and caught-between hazards. 
    Right: our lab setup studies a humanoid robot pushing an externally suspended payload toward a target region, requiring whole-body balance, contact-rich pushing, and accurate stopping. 
    Left image adapted from online material~\cite{youtubeHoistingExampleGXIMQnndCVU}.
    }
    \label{fig:1}
\end{center}

\section{Introduction}
Material handling often involves underactuated systems: payloads are not rigidly grasped or directly actuated, but are influenced through cables, hooks, slings, cranes, carts, or intermittent human contact. Such systems are common in construction, manufacturing, shipbuilding, logistics, and warehousing, where heavy payloads are transported by lifting equipment and then manually guided into their final positions. Hoisting is a representative example. Although cranes provide the primary lifting capability, final placement often still requires workers to stand near suspended loads and guide, stabilize, or push them by hand. Because suspended payloads can swing, respond with delay, and keep moving after contact, this stage creates struck-by and caught-between hazards~\cite{cdc2021liftzone,bls2023crane,niosh2006crane,oshaStruckBy}. Fig.~\ref{fig:1} illustrates both real hoisting operations and our lab setup, motivating the question: can humanoid robots safely operate around underactuated material-handling systems that are currently handled by human workers?


We study this question through \emph{humanoid hoisting}: using a humanoid robot to position an externally suspended payload through whole-body motion and contact-rich manipulation. Unlike standard pick-and-place or tabletop manipulation, the robot cannot directly control the payload pose or velocity; it can only influence the payload through base motion, arm motion, body posture, and contact forces. The payload therefore behaves like an underactuated pendulum, a structure widely studied in crane modeling and anti-sway control~\cite{abdel2003dynamics,ramli2017control,qiang2021anti,mojallizadeh2023modeling}. It may continue moving after the robot stops, overshoot the target, or retain residual oscillation. We therefore position humanoid hoisting as a new whole-body loco-manipulation problem, where success depends not only on feasible humanoid motion, but also on contact timing and stopping behavior for a dynamically active payload. VR teleoperation offers a practical way to collect demonstrations for initializing humanoid hoisting behaviors~\cite{zhang2018deep,rosen2018testing,han2023vr}. However, imitation learning does not directly optimize final placement accuracy and may accumulate errors during closed-loop execution~\cite{ross2011reduction}. Conversely, training from scratch with RL on a real humanoid is unsafe and sample-inefficient in this contact-rich setting~\cite{rajeswaran2017learning,wagenmaker2025steering}.

We present \textsc{HOIST}--\textbf{H}umanoid~\textbf{O}ptimized with~\textbf{I}mitation and \textbf{S}ample-efficient \textbf{T}uning for manipulating suspended loads. \textsc{HOIST} first uses VR whole body teleoperation collected demonstrations to finetune a high-level vision-language-action (VLA) policy, whose planner-level commands are executed by a whole-body controller. It then collects autonomous rollouts and applies iterative batched RL refinement with hoisting-specific rewards to improve placement accuracy and stopping behavior. This combines safe imitation initialization with task-level reward optimization, avoiding unsafe tabula-rasa exploration while addressing objectives not captured by imitation alone. Our contributions are: \begin{itemize} \item We introduce \emph{humanoid hoisting}, a whole-body loco-manipulation problem involving an externally suspended, indirectly driven, and oscillatory payload, opening a direction for humanoids to operate underactuated material-handling systems. 
\item We present \textsc{HOIST}, which combines VR teleoperation, supervised VLA finetuning, fixed whole-body execution, and sample-efficient RL refinement for suspended-payload manipulation. \item We evaluate suspended-payload positioning in simulation and on a real humanoid, comparing additional demonstrations with RL refinement and analyzing how rollout count and input modalities affect placement accuracy. \end{itemize}


\section{Related Work}
\paragraph{Robotic hoisting and suspended-payload control.}
Robotic systems have been widely studied for construction automation and material handling~\cite{bock2016construction,xu2020site}, where hoisting heavy payloads is common. 
Because suspended loads exhibit pendulum-like dynamics, prior work has studied modeling, trajectory planning, input shaping, and anti-sway control for crane and cable-suspended systems~\cite{abdel2003dynamics,ramli2017control,qiang2021anti,hoffman2020precision}. 
These methods typically control the lifting mechanism directly, whereas humanoid hoisting can only influence an externally suspended payload through intermittent whole-body contact. 
Our work therefore reformulates suspended-payload positioning as humanoid-assisted local manipulation, where the robot acts as an external contact-rich agent rather than part of the crane controller.

\paragraph{Humanoid whole-body control and loco-manipulation.}
Humanoid whole-body control has advanced from motion tracking to real humanoid systems with richer high-level interfaces~\cite{peng2018deepmimic,peng2018sfv,allshire2025visual,luo2025sonic,bjorck2025gr00t,gr00twbc2025}. 
Humanoid loco-manipulation further studies coordinated locomotion, balance, and contact-rich interaction through motion priors, residual policies, and compliance regulation~\cite{yoshida2007dexterous,zhang2025falcon,zhao2025resmimic,chen2025chip}. 
However, most prior tasks involve rigid or quasi-static objects whose motion is constrained by the robot or environment. 
In contrast, suspended payloads remain dynamically active after contact, making success depend on contact timing, accurate placement, and stopping behavior.

\paragraph{Reinforcement learning for robot control.}
Reinforcement learning can optimize robot behavior from task rewards and environment interaction, including high-dimensional contact-rich manipulation~\cite{rajeswaran2017learning,elguea2023review}. 
However, applying RL directly to humanoid hoisting is unsafe and sample-inefficient: early exploration may cause unstable motion, hard contacts, over-pushing, or large residual swing, motivating safety-aware and demonstration-guided learning~\cite{garcia2015comprehensive,gu2023human}. 
Our approach avoids tabula-rasa exploration by starting from a teleoperation-finetuned VLA policy and using RL to refine high-level commands from limited rollouts, optimizing placement and stopping while preserving safe demonstrated behavior~\cite{wagenmaker2025steering}.

\section{Methodology}
This section describes the technical formulation and implementation of \textsc{HOIST}. 
We first define the humanoid hoisting problem, then present the teleoperation data collection, high-level VLA policy, whole-body controller, and RL refinement procedure used to improve final payload placement.

\begin{figure}[t]
    \centering
    \includegraphics[width=\linewidth]{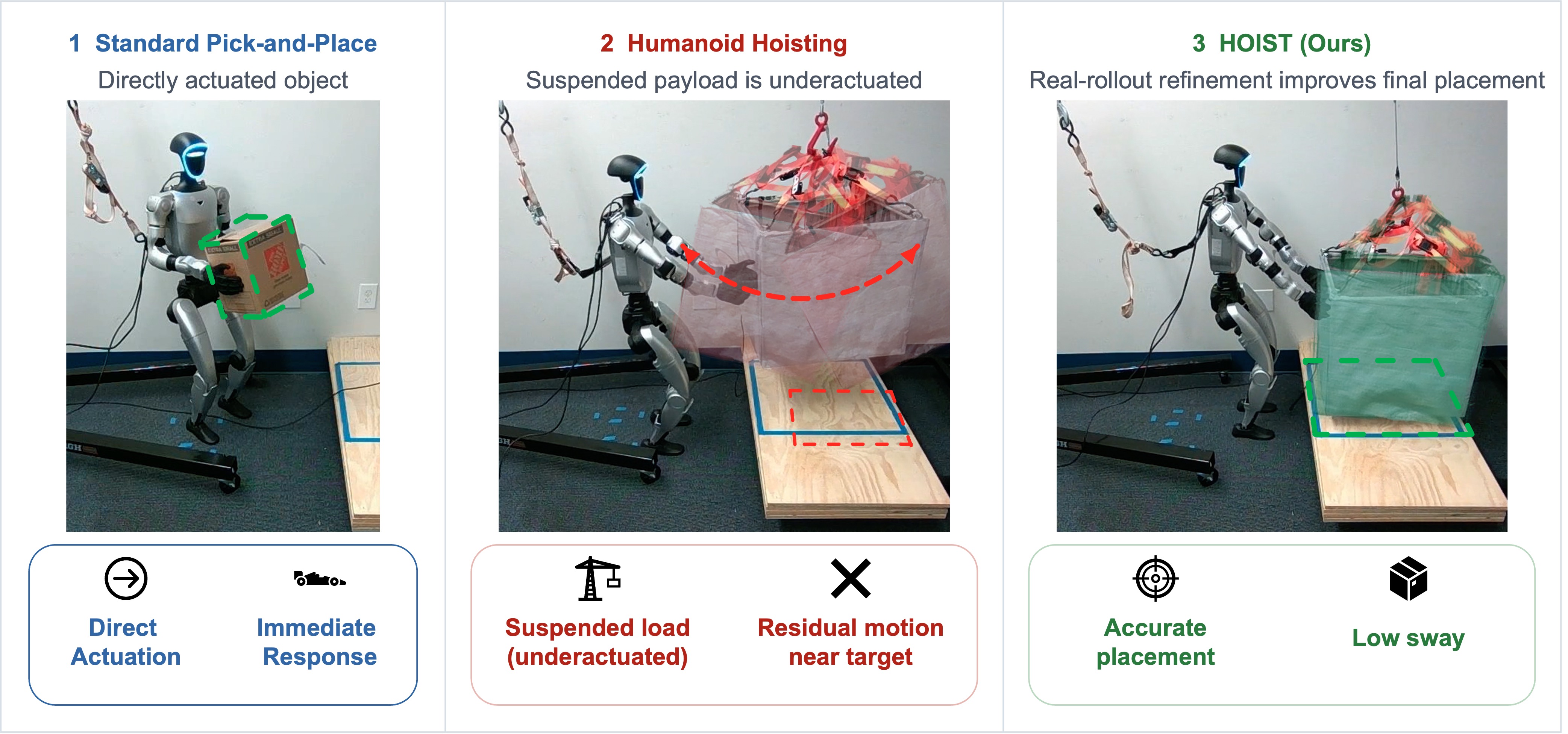}
     \vspace{-1.5em}
    \caption{Overview of humanoid hoisting. 
Left: in standard pick-and-place, the robot directly actuates the object and receives an immediate object response. 
Middle: in humanoid hoisting, the payload is externally suspended and underactuated, so the robot can only influence it through whole-body contact; this introduces delayed motion, overshoot, and residual swing near the target. 
Right: \textsc{HOIST} addresses this setting by learning high-level hoisting commands from teleoperation and refining them with iterative batched RL, improving final placement accuracy while reducing residual sway.}
    \label{fig:overview}
    \vspace{-1.0em}
\end{figure}

\begin{figure}[t]
    \centering
    \includegraphics[width=\linewidth]{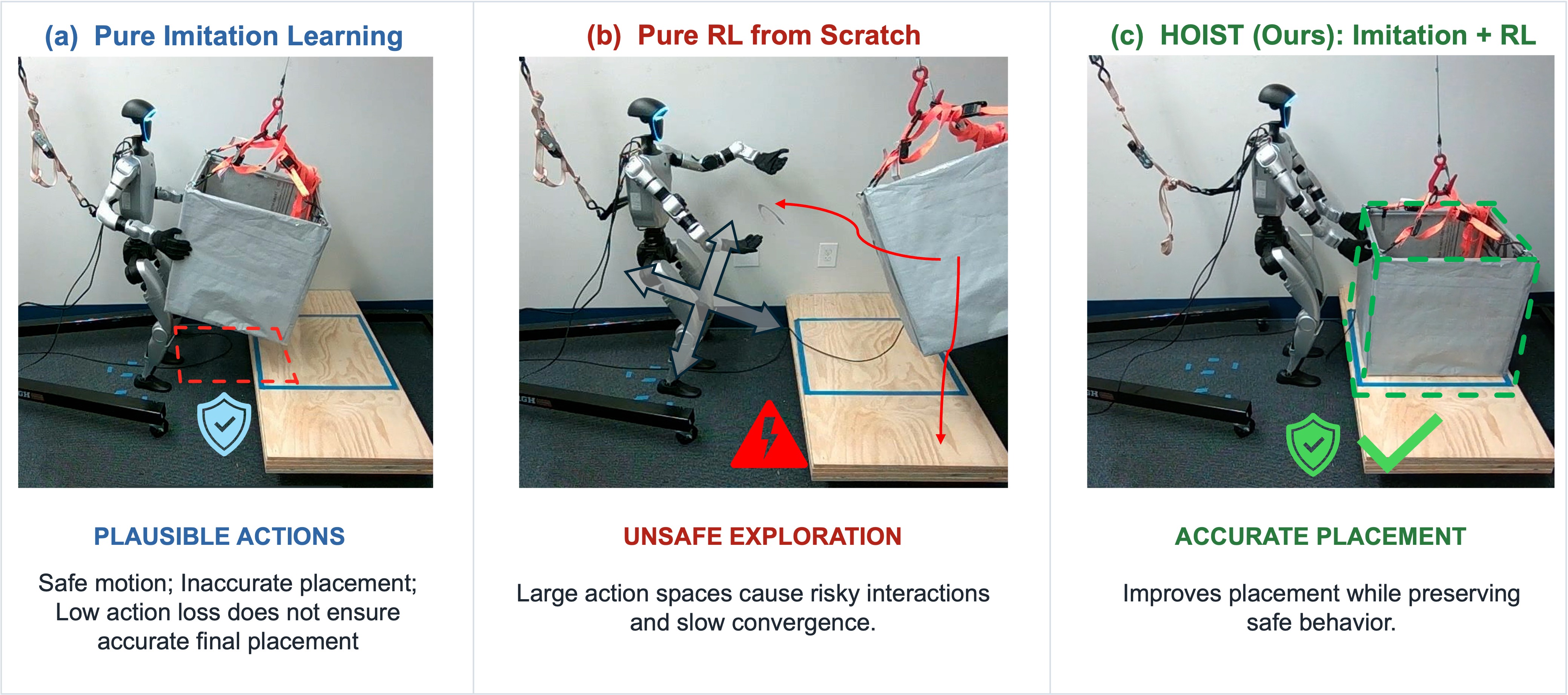}
    \vspace{-1.5em}
    \caption{Motivation for imitation-initialized RL. 
Pure imitation yields safe motions but may leave large placement errors, while pure RL from scratch is unsafe and hard to converge on real humanoids. 
\textsc{HOIST} starts from an imitation-learned policy and uses iterative batched RL refinement to improve task-level placement while preserving safe behavior.}
    \label{fig:learning-idea}
    \vspace{-1.5em}
\end{figure}

\subsection{Problem Formulation}
\label{sec:problem_formulation}

We consider \emph{humanoid hoisting}, where a humanoid robot assists the local positioning of an externally suspended payload. 
Given a task instruction and visual observations, the robot must approach the payload, establish physical interaction, push it toward a target region, and stop without inducing excessive residual swing. 
The robot does not actuate the crane or suspension mechanism. 
Instead, it can only influence the payload through whole-body motion and intermittent contact forces. 
Therefore, the task couples humanoid locomotion and balance, contact timing, payload displacement, and delayed pendulum-like swing.

We model humanoid hoisting as a task-conditioned, partially observed, closed-loop control problem. 
At each high-level decision step $t$, the policy receives the observation
\begin{equation}
    \mathbf{o}_t =
    \left\{
    \mathbf{I}^{\mathrm{ego,rgb}}_t,
    \mathbf{D}^{\mathrm{ego}}_t,
    \mathbf{I}^{\mathrm{side,rgb}}_t,
    \mathbf{x}^{r}_t,
    \ell,
    \mathbf{c}^{\mathrm{nav}}_{t-1}
    \right\},
\end{equation}
where $\mathbf{I}^{\mathrm{ego,rgb}}_t$ and $\mathbf{D}^{\mathrm{ego}}_t$ are the humanoid onboard RGB and depth observations, $\mathbf{I}^{\mathrm{side,rgb}}_t$ is an external side-view image that captures the payload-target relationship, $\mathbf{x}^{r}_t$ denotes the humanoid proprioceptive state, $\ell$ is the language instruction, and $\mathbf{c}^{\mathrm{nav}}_{t-1}$ is the previous navigation command. 
The proprioceptive state $\mathbf{x}^{r}_t$ include joint states and base/IMU information. 
We do not assume direct contact-force sensing or explicit payload-state input to the policy.

The high-level policy does not output joint torques. 
Instead, it predicts the next $H$ high-level planner-command increments:
\begin{equation}
    \Delta \mathbf{U}_t
    =
    \pi_{\theta}(\mathbf{o}_t)
    =
    \left(
    \Delta \mathbf{u}_{t},
    \Delta \mathbf{u}_{t+1},
    \ldots,
    \Delta \mathbf{u}_{t+H-1}
    \right),
\end{equation}
where $\pi_{\theta}$ is the learned high-level VLA policy with parameters $\theta$, $H$ is the action-chunk horizon, and $\Delta \mathbf{U}_t$ is the predicted command-increment chunk at time $t$. 

Each single-step command increment $\Delta \mathbf{u}_{\tau}$ updates the head target, left-hand target, right-hand target, navigation command, and desired base height:
\begin{equation}
    \Delta \mathbf{u}_{\tau}
    =
    \left[
    \Delta \mathbf{y}^{\mathrm{head}}_{\tau},
    \Delta \mathbf{y}^{L}_{\tau},
    \Delta \mathbf{y}^{R}_{\tau},
    \Delta \mathbf{c}^{\mathrm{nav}}_{\tau},
    \Delta h^{\mathrm{base}}_{\tau}
    \right],
    \quad \tau=t,\ldots,t+H-1.
\end{equation}
Here, $\tau$ indexes a future high-level step within the predicted command chunk, $\mathbf{y}^{\mathrm{head}}_{\tau}$ is the head tracking target, $\mathbf{y}^{L}_{\tau}$ and $\mathbf{y}^{R}_{\tau}$ are the left- and right-hand tracking targets, $\mathbf{c}^{\mathrm{nav}}_{\tau}$ is the navigation command, and $h^{\mathrm{base}}_{\tau}$ is the desired base-height command.

The corresponding full planner command at time $t$ is
\begin{equation}
    \mathbf{u}_{t}
    =
    \left[
    \mathbf{y}^{\mathrm{head}}_{t},
    \mathbf{y}^{L}_{t},
    \mathbf{y}^{R}_{t},
    \mathbf{c}^{\mathrm{nav}}_{t},
    h^{\mathrm{base}}_{t}
    \right].
\end{equation}
In words, the VLA policy only decides short-horizon task-level motions, such as where the hands and body should move next; the low-level controller converts these commands into balanced whole-body motion.

The execution stack maintains the previous full planner command $\mathbf{u}_{t-1}$, while the policy receives only the previous navigation command $\mathbf{c}^{\mathrm{nav}}_{t-1}$ as an explicit command-history input. 
At execution time, we apply the first increment in the chunk to update the current planner command:
\begin{equation}
    \mathbf{u}_t = \mathbf{u}_{t-1} + \Delta \mathbf{u}_t.
\end{equation}
The updated command is then executed by a humanoid whole-body control stack:
\begin{equation}
    \mathbf{a}^{\mathrm{motor}}_t
    =
    \pi_{\mathrm{wbc}}
    \left(
    \mathbf{x}^{r}_t,
    \mathbf{u}_t
    \right),
\end{equation}
where $\pi_{\mathrm{wbc}}$ is a fixed decoupled whole-body controller~\cite{luo2025sonic}, and $\mathbf{a}^{\mathrm{motor}}_t$ denotes the low-level motor command. 
This hierarchy lets the learned policy operate at the task-command level, while whole-body tracking, balance, and joint-level stabilization are handled by the fixed execution stack.

\begin{figure}[t]
    \centering
     \includegraphics[width=\linewidth]{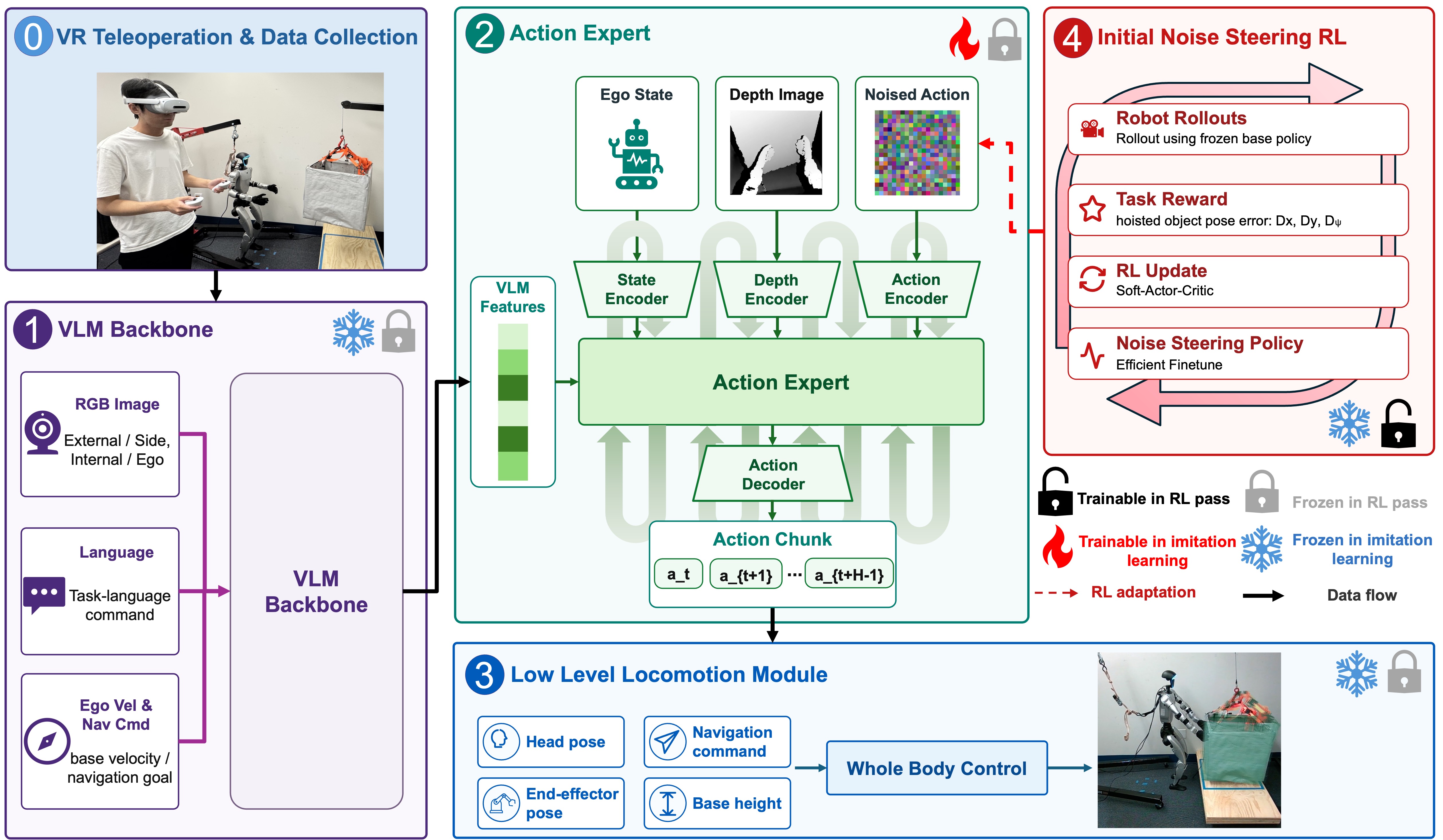}
\caption{
Overview of the \textsc{HOIST} pipeline. 
VR teleoperation provides hoisting demonstrations to finetune a high-level VLA policy, which predicts short-horizon planner-command chunks. 
A fixed whole-body controller executes these commands as stable humanoid motion. 
Autonomous VLA rollouts are then used for iterative batched RL with hoisting-specific rewards to improve placement accuracy, residual swing, and stopping behavior.
}
    \label{fig:pipeline}
    \vspace{-1.0em}
\end{figure}

\subsection{Overview of HOIST}
\label{sec:overview} 
Our method consists of three stages: teleoperation data collection, supervised VLA finetuning, and RL refinement. 
First, a human operator uses a VR headset and controllers to teleoperate the humanoid robot through the hoisting task. 
Second, we use the collected teleoperation data to finetune A high-level VLA policy, so that the humanoid can autonomously generate hoisting-related planner commands from visual observations and task instructions. 
Third, we deploy the finetuned VLA policy to collect autonomous rollouts and use these rollouts for iterative batched flow-matching-steering RL~\cite{wagenmaker2025steering} refinement.

\paragraph{Teleoperation data collection.}
We collect demonstrations using a PICO VR headset and controllers. 
The operator guides the humanoid through the full hoisting sequence: approach, contact, push, and stop. 
Each trajectory is stored as
\begin{equation}
    \tau_i =
    \left(
    \left\{
    \mathbf{o}_t,
    \mathbf{u}^{\mathrm{demo}}_t,
    \mathbf{a}^{\mathrm{motor}}_t
    \right\}_{t=1}^{T_i},
    y_i
    \right),
\end{equation}
where $\mathbf{u}^{\mathrm{demo}}_t$ is the planner command produced during teleoperation and $y_i$ denotes trajectory-level outcomes such as final placement error and residual motion.

We construct the supervised action-chunk target as
\begin{equation}
    \Delta \mathbf{U}^{\mathrm{demo}}_t
    =
    \left(
    \Delta \mathbf{u}^{\mathrm{demo}}_{t},
    \ldots,
    \Delta \mathbf{u}^{\mathrm{demo}}_{t+H-1}
    \right),
    \qquad
    \Delta \mathbf{u}^{\mathrm{demo}}_{\tau}
    =
    \mathbf{u}^{\mathrm{demo}}_{\tau}
    -
    \mathbf{u}^{\mathrm{demo}}_{\tau-1}.
\end{equation}
These demonstrations initialize the high-level policy with safe approach, pushing, and stopping behavior.

\paragraph{High-level VLA policy.}
We build the high-level policy by modifying GR00T N1.6~\cite{bjorck2025gr00t} for humanoid hoisting. 
The policy takes multimodal observations and task instructions as input and outputs a short sequence of relative commands for both hands, the navigation command, the head pose, and the base height. 
To make the policy better aware of the robot's current motion, we add the navigation command as an additional modality to the vision-language model. 
We also add depth images and robot ego-velocity to the flow-matching based action expert, so that the action expert can condition on both scene geometry and the humanoid's current physical state. 
During supervised finetuning, most VLM layers are frozen; we finetune only the final four layers of VLM, the diffusion-transformer action expert, and the encoders for the newly introduced modalities. 
This design adapts the foundation policy to humanoid hoisting while preserving the general visual-language and action priors learned by the pretrained model.

\paragraph{Humanoid whole-body controller.}
We use the GR00T Whole-Body Control deployment stack as a fixed execution layer. 
The high-level VLA policy outputs planner-level command updates, which are accumulated into inputs as target velocities, hand/head tracking targets, and base height. 
We do not train or finetune the planner or low-level controller for the hoisting task. 
This design keeps the low-level motion prior fixed and shifts task adaptation to the high-level VLA policy and RL refinement module. 
Thus, \textsc{HOIST} learns how to command hoisting-relevant motion without learning joint-level actuation or modifying the whole-body controller.

\paragraph{Reinforcement-learning refinement.}
After supervised finetuning, we refine the high-level command generator using VLA rollouts. 
In each domain, we deploy the finetuned VLA policy together with the fixed whole-body controller to collect a small batch of autonomous hoisting rollouts. 
These rollouts are added to a refinement dataset and labeled with hoisting-specific rewards. In our current implementation, the reward is a simple weighted combination of translational placement errors and the raw angular error, rather than a learned or heavily shaped reward model.
We then perform an offline actor-critic update on the accumulated rollout dataset. 
The updated refinement module is redeployed to collect the next batch of rollouts, and the same collect-train cycle is repeated. 

During rollouts, we record the initial noise used by the flow-matching action expert and train an actor-critic module to predict which initial-noise vector should be used for a given state. 
The actor outputs this steering vector deterministically rather than predicting a mean and variance and then sampling from them. 
Throughout RL, the VLA policy itself remains frozen; only the noise-steering actor-critic module is updated. 
For efficiency, we store VLM features from the frozen policy and reuse them to avoid repeated encoding of the same visual-language observations. 
This batched flywheel allows the refinement stage to improve placement accuracy with a small number of task rollouts while preserving the behavior learned from teleoperation.

\subsection{Metrics}
\label{sec:evaluation_metrics}

Let $(x_T, y_T, \psi_T)$ denote the final planar pose of the payload center and $(x^{\star}, y^{\star}, \psi^{\star})$ denote the target pose. We evaluate placement accuracy using absolute translational and yaw errors:
\begin{equation}
\mathbf{e}_{\mathrm{place}}
=
\begin{bmatrix}
\Delta x \
\Delta y \
\Delta \psi
\end{bmatrix}
=
\begin{bmatrix}
\left|x_T - x^{\star}\right| \
\left|y_T - y^{\star}\right| \
\left|\mathrm{wrap}(\psi_T - \psi^{\star})\right|
\end{bmatrix}.
\end{equation}
Here, $\Delta x$ and $\Delta y$ measure the terminal center offset of the payload along the $x$- and $y$-directions, respectively, while $\Delta \psi$ measures its terminal rotational error about the vertical $z$-axis. In the result tables, we report $\Delta x$, $\Delta y$, $\Delta \psi$, and the Manhattan terminal position error $\Delta x + \Delta y$.


\section{Experiments and Results}

\subsection{Experiment Setup}

We evaluate \textsc{HOIST} in simulation and on a real humanoid platform. 
Each trial starts with the payload away from the target; the humanoid must position it through contact without actuating the suspension mechanism. 
In each domain, VLA-50 is finetuned on $50$ teleoperated demonstrations, VLA-80 uses $80$ demonstrations, and \textsc{HOIST} starts from VLA-50 before applying up to $30$ same-domain RL rollouts. 
All methods use the same frozen GR00T whole-body execution stack, isolating changes to the high-level command policy.

\subsection{Reinforcement-Learning Progression}

Table~\ref{tab:rollout_progression} reports placement accuracy after each batch of RL rollouts. 
The resulting trend shows whether same-domain refinement steadily improves the VLA-50 initialization. A decreasing error trend indicates that rollout feedback corrects placement bias left by imitation learning.


\begin{table}[H]
    \centering
    \small
    \vspace{-1em}
    \caption{
    Placement accuracy with increasing numbers of RL rollouts.
    }
    \label{tab:rollout_progression}
    \resizebox{\linewidth}{!}{%
    \begin{tabular}{llcccc}
        \toprule
        Domain & Method 
        & $\Delta x$(cm)$\downarrow$ 
        & $\Delta y$ (cm)$\downarrow$ 
        & $\Delta \psi$ (deg)$\downarrow$ 
        & $|\Delta x| + |\Delta y|$ (cm)$\downarrow$ \\
        \midrule
        Simulation & Human Expert Teleop & 4.35 & 3.00 & 8.48 & 7.35 \\
        Simulation & VLA-50 & 21.56 & 0.88 & 3.85 & 22.44 \\
        Simulation & VLA-50 + 10 RL rollouts & 15.09 & 1.83 & 4.58 & 16.92 \\
        Simulation & VLA-50 + 20 RL rollouts & \best{1.72} & \best{0.82} & \best{0.29} & \best{2.54} \\
        Simulation & VLA-50 + 30 RL rollouts & 5.16 & 1.09 & 2.76 & 6.25 \\
        \midrule
        Real platform & VLA-50 & \best{1.60} & 7.68 & 14.5 & 9.28 \\
        Real platform & VLA-50 + 10 RL rollouts & 2.15 & 5.22 & 21.0 & 7.37 \\
        Real platform & VLA-50 + 20 RL rollouts & 4.20 & \best{2.49} & \best{12.1} & 6.69 \\
        Real platform & VLA-50 + 30 RL rollouts & 1.64 & 4.74 & 28.9 & \best{6.38} \\
        \bottomrule
    \end{tabular}}
    \vspace{-1em}
\end{table}

\subsection{Reinforcement Learning versus Additional Demonstrations}

Table~\ref{tab:rl_vs_demos} compares additional demonstrations with same-domain RL rollouts after the initial $50$ demonstrations.  \textsc{HOIST} achieves better performance than VLA-80, showing that optimizing closed-loop placement rewards is more effective than adding demonstrations alone. 
This supports using RL to correct final-position errors that remain after imitation learning.


\begin{table}[H]
    \centering
    \small
    \vspace{-1em}
    \caption{Additional demonstrations versus RL refinement.}
    \label{tab:rl_vs_demos}
    \resizebox{\linewidth}{!}{%
    \begin{tabular}{llcccccc}
        \toprule
        Domain & Method & Demos & RL rollouts 
        & $\Delta x$ (cm)$\downarrow$ 
        & $\Delta y$ (cm)$\downarrow$ 
        & $\Delta \psi$ (deg) 
        & $|\Delta x| + |\Delta y|$ (cm)$\downarrow$ \\
        \midrule
        Simulation & VLA-50 & 50 & 0 & 21.56 & \best{0.88} & 3.85 & 22.44  \\
        Simulation & VLA-80 & 80 & 0 & 15.69 & 2.89 & 12.14 & 18.58 \\
        Simulation & \textsc{HOIST (Ours)} & 50 & 30 & \best{5.16} & 1.09 & \best{2.76} & \best{6.25} \\
        \midrule
        Real platform & VLA-50 & 50 & 0 & 1.60 & 7.68 & \best{14.5} & 9.28  \\
        Real platform & VLA-80 & 80 & 0 & \best{0.54} & 8.03 & 27.0 & 8.57 \\
        Real platform & \textsc{HOIST (Ours)} & 50 & 30 & 1.64 & \best{4.74} & 28.9 & \best{6.38} \\
        \bottomrule
    \end{tabular}}
    \vspace{-1.0em}
\end{table}

\subsection{Observation-Modality Ablation}

Table~\ref{tab:modality_ablation} evaluates the VLA policy before RL refinement under different input modalities. 
The goal is to measure whether depth and navigation-command conditioning improve the imitation-learned initialization. All rows use the same number of demonstrations and no RL rollouts.


\begin{table}[H]
    \centering
    \small
    \vspace{-1em}
    \caption{Ablation of VLA input modalities before RL refinement.}
    \label{tab:modality_ablation}
    \resizebox{\linewidth}{!}{%
    \begin{tabular}{lcccccc}
        \toprule
        VLA variant & Depth & Nav command 
        & $\Delta x$ (cm)$\downarrow$ 
        & $\Delta y$ (cm)$\downarrow$ 
        & $\Delta \psi$ (deg)$\downarrow$ 
        & $|\Delta x| + |\Delta y|$ (cm)$\downarrow$ \\
        \midrule
        Base VLA & -- & -- & 1.54 & 10.52 & 26.6 & 12.06  \\
        + Depth & \checkmark & -- & \best{0.78} & 8.78 & 14.8 & 9.57 \\
        + Nav command & -- & \checkmark & 1.30 & 9.90 & 32.5 & 11.20 \\
        + Nav command + Depth & \checkmark & \checkmark & 1.60 & \best{7.68} & \best{14.5} & \best{9.28} \\
        \bottomrule
    \end{tabular}}
    \vspace{-1em}
\end{table}

Depth improves contact-geometry awareness, while navigation-command conditioning improves the use of current base motion. 
The combined variant performs best, supporting the use of both modalities before RL refinement.

\subsection{Residual Motion Analysis}
\begin{figure}[H]
    \centering
    \includegraphics[width=0.7\linewidth]{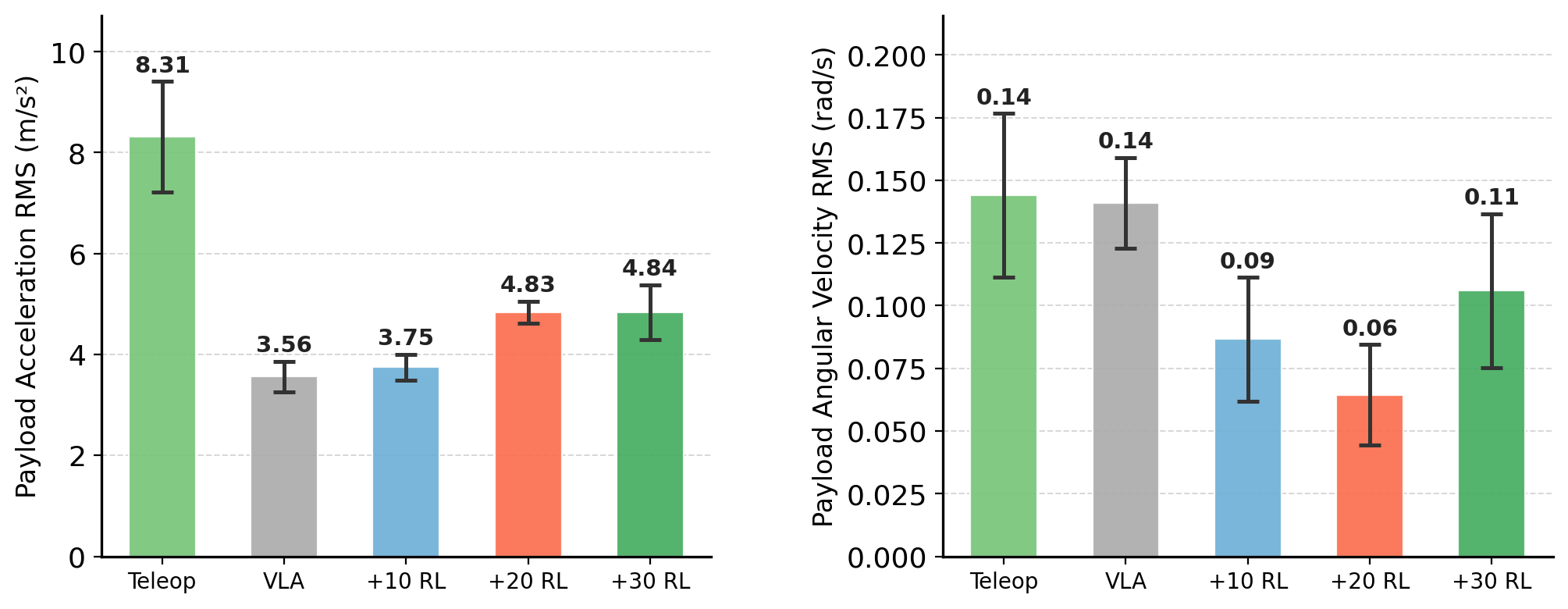}
    \caption{Residual-motion analysis from an IMU mounted on the suspended payload.}
    \label{fig:imu-residual-motion}
        \vspace{-1.0em}
\end{figure}
We further analyze payload dynamics throughout the lifting process using an IMU mounted on the suspended load. As shown in Fig.~\ref{fig:imu-residual-motion}, compared with the imitation-only policy, HOIST yields higher accelerometer readings but reduces gyroscope responses by approximately half, suggesting reduced angular motion with more active corrective adjustments during lifting. Meanwhile, both learned policies exhibit lower accelerometer and gyroscope readings than human teleoperation demonstrations, which may reflect the more frequent closed-loop corrections performed by human operators.


\section{Discussion}
The improvement from 30 RL rollouts suggests that the imitation-learned VLA policy has systematic deployment bias. Supervised finetuning matches expert actions on the demonstration distribution, but the learned policy is not executed under that same distribution: once deployed, its own prediction errors affect contact timing, pushing strength, and stopping behavior, producing a different closed-loop rollout distribution. Thus, the policy may consistently under-push, over-push, or stop at a biased terminal position.

Additional demonstrations add more expert-distribution data, but they do not directly identify this deployment-time bias. In contrast, autonomous rollouts reveal the biased trajectories generated by the current VLA policy itself. HOIST can then use task-level rewards to learn an initial-noise correction that steers the frozen policy away from these systematic errors. This makes RL rollouts more useful than extra demonstrations for correcting final placement accuracy.

\section{Limitation}

The primary limitation of the current system lies in the low-level whole-body controller, which has not been fine-tuned for interaction with suspended payloads. Specifically, the policy can only influence robot execution through a limited interface consisting of navigation commands, head pose, base height, and bimanual hand poses, without explicitly adapting the applied force to different payload weights. Consequently, the system struggles to generate sufficiently large and adaptive forces when manipulating heavier objects, limiting its ability to handle heavy payloads.

\section{Conclusion}

We presented \textsc{HOIST}, which initializes humanoid hoisting from teleoperated demonstrations and improves final placement through rollout-level RL while keeping the whole-body controller fixed. 
The learned initial-noise correction changes command sequences, which propagate through humanoid-payload contact dynamics to reduce final placement error and residual swing. 
Future work will expand payload and suspension conditions, add explicit safety constraints, and study coordinated humanoid-hoist control.

\clearpage
\acknowledgments{If a paper is accepted, the final camera-ready version will (and probably should) include acknowledgments. All acknowledgments go at the end of the paper, including thanks to reviewers who gave useful comments, to colleagues who contributed to the ideas, and to funding agencies and corporate sponsors that provided financial support.}


\bibliography{example}  
\end{document}